\pdfoutput=1
\documentclass[10pt,twocolumn,letterpaper]{article}
\usepackage{amssymb}
\usepackage{pifont}
\usepackage{bbm}
\usepackage{placeins}
\usepackage{graphicx}
\usepackage{amsmath}
\usepackage[accsupp]{axessibility}

\DeclareMathOperator*{\argmax}{\arg\,\max}

\newcommand{\myparagraph}[1]{\vspace{0.4em}\textbf{#1}}

\def\cvprPaperID{3557}
\def\confName{CVPR}
\def\confYear{2023}

\def\paperTitle{Zero-guidance Segmentation Using Zero Segment Labels}

\def\authorBlock{
    \fontsize{10.5}{7}\selectfont Pitchaporn Rewatbowornwong\thanks{Equal contribution}\:\,$\vcenter{\hbox{\includegraphics[width=0.005\textwidth\vspace*{0.35em}]{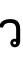}}}$ \qquad
    Nattanat Chatthee\footnotemark[1]\:\,$\vcenter{\hbox{\includegraphics[width=0.005\textwidth\vspace*{0.35em}]{figs/vistec.pdf}}}$\qquad
    Ekapol Chuangsuwanich$\vcenter{\hbox{\includegraphics[width=0.008\textwidth\vspace*{0.35em}]{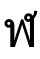}}}$ \qquad
    Supasorn Suwajanakorn$\vcenter{\hbox{\includegraphics[width=0.005\textwidth\vspace*{0.35em}]{figs/vistec.pdf}}}$\\
     \fontsize{10.5}{7}\selectfont$\vcenter{\hbox{\includegraphics[width=0.005\textwidth\vspace*{0.35em}]{figs/vistec.pdf}}}$\hspace{1pt}VISTEC, Thailand \qquad \qquad \qquad
    $\vcenter{\hbox{\includegraphics[width=0.008\textwidth\vspace*{0.35em}]{figs/chula.pdf}}}$\hspace{1pt}Chulalongkorn University, Thailand\\
    {\tt\small \{pitchaporn.r\_s18, nattanatc\_pro, supasorn.s\}@vistec.ac.th \qquad 
    ekapol.c@chula.ac.th}
    
}

\newif\ifreview 
\newif\ifarxiv \newcommand{\arxiv}{\arxivtrue}
\newif\ifcamera 
\newif\ifrebuttal 

\arxiv 

\ifreview \usepackage[review]{cvpr} \fi
\ifarxiv \usepackage[pagenumbers]{cvpr} \fi
\ifrebuttal \usepackage[rebuttal]{cvpr} \fi
\ifcamera \usepackage{cvpr} \fi

\usepackage{graphicx}
\usepackage{amsmath}
\usepackage{amssymb}
\usepackage{booktabs}
\usepackage{mathrsfs}

\usepackage{times}
\usepackage{microtype}
\usepackage{epsfig}
\usepackage[table,xcdraw,dvipsnames]{xcolor}
\usepackage{caption}
\usepackage{float}
\usepackage{placeins}
\usepackage{color, colortbl}
\usepackage{stfloats}
\usepackage{enumitem}
\usepackage{tabularx}
\usepackage{xstring}
\usepackage{multirow}
\usepackage{xspace}
\usepackage{url}
\usepackage{subcaption}
\usepackage{xcolor}
\usepackage[hang,flushmargin]{footmisc}

% ...

\ifarxiv  \fi

\newcommand{\R}[1]{{%
    \textbf{%
        \ifstrequal{#1}{1}{\textcolor{red}{R#1}}{%
        \ifstrequal{#1}{2}{\textcolor{blue}{R#1}}{%
        \ifstrequal{#1}{3}{\textcolor{magenta}{R#1}}{%
        \ifstrequal{#1}{4}{\textcolor{teal}{R#1}}{%
                           \textcolor{cyan}{R#1}%
        }}}}%
    }%
}}

\usepackage{xr-hyper}

\makeatletter
\newcommand*{\addFileDependency}[1]{
  \typeout{(#1)}
  \@addtofilelist{#1}
  \IfFileExists{#1}{}{\typeout{No file #1.}}
}

\makeatother

\usepackage[pagebackref,breaklinks,colorlinks]{hyperref}
\usepackage[capitalize]{cleveref}
\crefname{section}{Sec.}{Secs.}
\crefname{table}{Table}{Tables}
\crefname{figure}{Fig.}{Figs.}

\begin{document}

\title{\paperTitle}
\author{\authorBlock}
\maketitle
\begin{abstract}
The joint visual-language model CLIP has enabled new and exciting applications, such as open-vocabulary segmentation, which can locate any segment given an arbitrary text query. 
In our research, we ask whether it is possible to discover semantic segments without any user guidance in the form of text queries or predefined classes, and label them using natural language automatically?
We propose a novel problem \textbf{zero-guidance segmentation} and the first baseline that leverages two pre-trained generalist models, DINO and CLIP, to solve this problem without any fine-tuning or segmentation dataset.
The general idea is to first segment an image into small over-segments, encode them into CLIP's visual-language space, translate them into text labels, and merge semantically similar segments together. The key challenge, however, is how to encode a visual segment into a segment-specific embedding that balances global and local context information, both useful for recognition.
Our main contribution is a novel attention-masking technique that balances the two contexts by analyzing the attention layers inside CLIP. We also introduce several metrics for the evaluation of this new task. 
With CLIP's innate knowledge, our method can precisely locate the Mona Lisa painting among a museum crowd (Figure \ref{fig:teasor}). More results are available at \emph{\small \url{https://zero-guide-seg.github.io/}}.
\end{abstract}
\section{Introduction}
\label{sec:intro}
Semantic segmentation is a core computer vision problem that seeks to partition an image into semantic regions. 
Traditionally, the semantic classes of interest need to be predefined and are limited in number \cite{9356353}. Earlier methods thus cannot generalize beyond the training classes. 
With recent advances in joint vision-language representation learning, e.g., CLIP \cite{radford2021learning}, newer methods \cite{liang2022open, li2022language, xu2021simple} can successfully predict segments corresponding to arbitrary text queries in a novel task called open-vocabulary segmentation. These segmentation methods are guided by a text query, which describes what already exists in the image and must be provided by the user.
Another meaningful milestone, however, is how we can segment an image \emph{without} user input or guidance like text queries or predefined classes, and label such segments automatically using natural language. Our work provides the first baseline for this novel problem, referred to as \emph{zero-guidance segmentation\footnote{We avoid calling this \emph{zero-shot} segmentation because 
it has been used in the literature for setups that \emph{require} a text query but allow it to be unseen during training, whereas our problem does not require any text query.}.}   

\begin{figure}
  \includegraphics[scale=0.59]{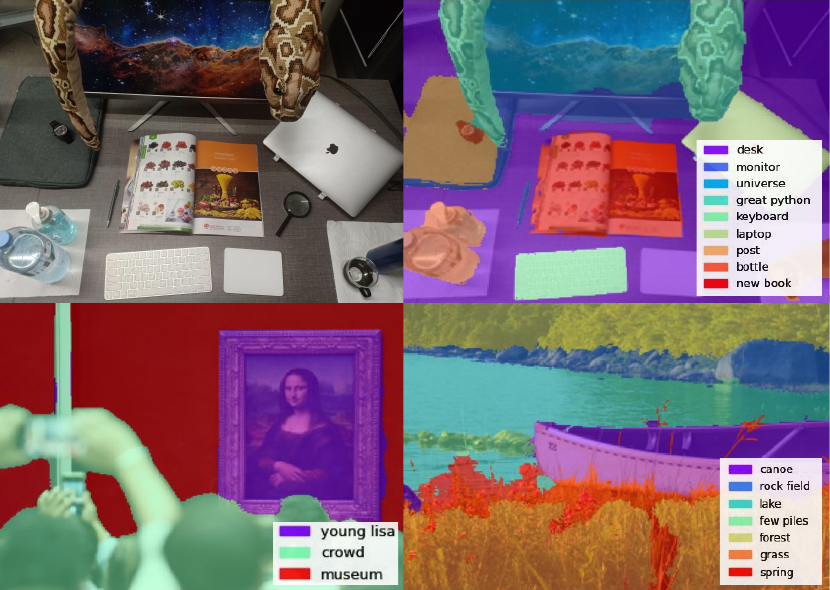}
  \caption{\textbf{Zero-guidance Segmentation} segments input images and generated text labels for all segments without any guidance or prompting. Our method produces these results using only pretrained networks with no fine-tuning or annotations.}
  \label{fig:teasor}

  \vspace{-1.em}
\end{figure}

Our work is inspired by a recent research direction that solves segmentation by leveraging CLIP \cite{xu2021simple, zhou2021denseclip};
however, our key distinction is that we require no segmentation datasets, no text query guidance, and no additional training or fine-tuning. This problem is challenging partly because CLIP has been trained with image captions that globally describe the scenes and provide no spatially specific information for learning segmentation.
Surprisingly, we show that it is possible to distill the learned knowledge from two generalist models: a self-supervised visual model, DINO \cite{caron2021emerging}, and a visual-language model, CLIP \cite{radford2021learning},
to solve zero-guidance segmentation without further training.

The overall idea is to first over-segment an image into small segment candidates, then translate each segment into words, and finally join semantically similar segments to form the output segments. 
In particular, we identify segment candidates by clustering deep pixel-wise features from a DINO that takes as input our image. Despite using no training labels, DINO has been shown to produce class-discriminative features, allowing unsupervised segmentation of the primary object in an image \cite{caron2021emerging} or part co-segmentation across images \cite{amir2021deep}. However, our main challenge lies in the next step, which maps each segment into a meaningful representation that can be later translated to words. We leverage CLIP's joint space to model this intermediate representation.


A naive way to project a segment to CLIP's joint space is to input the segment directly into CLIP's image encoder, but this entirely ignores the surrounding context needed for object recognition and disambiguation \cite{zhao2017pyramid,chen2017rethinking,article}.
Alternatively, masking can be applied inside the attention layers, as done with a transformer-based segmentation network \cite{cheng2022masked}.
However, when applied to CLIP's encoder, which was not trained for segmentation,
these techniques struggle to produce segment-specific embeddings due to the domination of global context information.
Evidence in \cite{zhou2021deepvit, xue2022deeper, gong2021vision} also suggests that a transformer trained with image-level annotations, such as CLIP, may lose local information in its tokens in later layers. 
We discovered similar issues: masking in earlier layers removes global contexts and hurts recognition, whereas masking in later layers fails to focus the embedding on a given segment, resulting in all embeddings describing the same dominant object in the image.




Another difficulty in balancing global and local contexts
is that different objects may require different degrees of context balancing.
For small objects, their CLIP embeddings can be dominated by global contexts, which describe other prominent objects in the scene. This phenomenon matches the characteristics of CLIP's training captions, which often ignore unimportant objects in the image. As a result, less prominent objects may require less of the global contexts to highlight their semantics and local contexts.


To solve this, we introduce a novel attention-masking technique called \emph{global subtraction}, which helps adjust the influence of global contexts in the output embedding. The key idea is to first estimate the saliency or the presence of a given segment in the global contexts by analyzing CLIP's attention values. Then, this saliency value will be used to determine how much global contexts should be attenuated in the segment's embedding. The resulting embedding in CLIP's joint vision-language space allows us to readily translate it to text labels with an existing image-to-text generation algorithm \cite{tewel2022zerocap}. And finally, we merge semantically similar segments with simple thresholding by considering both their visual and text similarities.


To evaluate our algorithm that can output arbitrary text labels, we also propose new evaluation metrics. Evaluating an algorithm under this setup is not straightforward as predicted labels may not necessarily match predefined labels in the test set but can still be correct.
This may result from the use of synonyms, such as ``cat'' vs. ``feline,'' or differences in label granularity, such as ``cat'' vs. ``orange cat'' or ``cat's nose'' or ``kitten.'' Generally, there is no single correct level of granularity, and each dataset may arbitrarily adopt any level. To address this, we propose to first map the predicted semantic labels to the existing ones in a given test set. After that, we can use standard measurements, such as segment IoU, to evaluate the results as if the algorithm performs segmentation with the predefined test classes. We also introduce Segment Recall, which measures how often ground-truth objects are discovered, and Text Generation Quality, which tests the quality of our embedding technique given ground-truth oracle segmentation.

Our technique can automatically segment an image into meaningful segments as shown in Figure \ref{fig:teasor} without any supervision or text guidance. There are still performance gaps between our technique and other supervised methods or methods fine-tuned on segmentation datasets---but none can specifically solve our problem that lacks user guidance. Nonetheless, we provide a detailed analysis on obstacles that lie ahead as well as ablation studies for the first approach to this problem. In summary, our contributions are:



\begin{itemize}
\vspace{-.2em}
\item We introduce the first baseline to a novel problem, \emph{zero-guidance segmentation}, which aims to segment and label an input image in natural language without predefined classes or text query guidance. Our method does not require a segmentation dataset or fine-tuning.
\item We propose a novel attention-masking technique to convert a segment into an embedding in CLIP's joint space by balancing global and local contexts.
\item We present evaluation metrics for the proposed setup.

\end{itemize}

\section{Related Work}
\label{sec:related}
\textbf{Open-vocabulary segmentation.}
This problem aims to predict segments in an input image that correspond to a set of input texts not necessarily seen during training. 
Prior solutions often involve a shared latent space between the image and text domains. 
OpenSeg \cite{ghiasi2021open} uses datasets of images, captions, and class-agnostic segmentation masks to train a mask proposal network before matching the predicted masks with nouns in the captions using a shared latent space. 
OVSeg \cite{liang2022open} built a segmentation pipeline that fine-tunes CLIP on masked images to make it more suitable for masked image classification. 
Xu et al. \cite{xu2021simple} proposed a zero-shot segmentation baseline by matching CLIP's embeddings of masked images to text embeddings of classes.
ZegFormer \cite{ding2022decoupling} performs class-agnostic pixel grouping to create segments and uses CLIP to classify them. 
Lseg \cite{li2022language} trains an image pixel encoder that encodes each pixel into an embedding that is close to the corresponding text labels' embeddings in CLIP's space.
These methods show impressive results but still demand expensive segmentation labels.

To avoid the use of segmentation datasets, GroupViT \cite{xu2022groupvit} proposes a new method based on hierarchical vision transformers where the visual tokens represent arbitrary regions instead of patches in a square grid. By using only image-caption pairs, GroupVit can match each region from its visual tokens to input text prompts.
Zhou et al. \cite{zhou2021denseclip} modifies CLIP for text-guided segmentation and employs self-training to improve the results.
In contrast, our work requires neither additional training nor text prompts but can discover semantic segments and label them automatically.

\textbf{Attention masking in transformer.}
Masking self-attention is a common practice in NLP to input a word sequence more efficiently \cite{vaswani2017attention}.
In computer vision, few explorations exist:
Mask2Former \cite{cheng2022masked} solves supervised segmentation 
by masking self-attention layers of a transformer decoder, achieving state-of-the-art results. 
Unlike Mask2Former, which is trained on specific segmentation datasets, our method and the base models we used (CLIP and DINO-ViT) do not have any explicit segmentation supervision. We found that using the masking mechanism of Mask2Former yields noisy CLIP embeddings, which are often heavily biased toward the foreground objects. This can be solved by our proposed global subtraction technique.


\textbf{Image segmentation with DINO.}
DINO \cite{caron2021emerging} is a model that uses self-distillation to learn rich features of an input image with no supervision and has been used as a pre-trained network or representation extractor in many tasks \cite{vaze2022gcd, wang2022tokencut, Simeoni2021LocalizingOW, hamilton2022unsupervised}.
Caron et al. \cite{caron2021emerging} demonstrated that DINO's features effectively capture object boundaries and scene layout \cite{caron2021emerging}, and Hamilton et al. \cite{hamilton2022unsupervised} further showed that these features can perform segmentation of not only foreground objects but also other elements in the background, such as the sky.
Our method uses a simple DINO-based clustering, inspired by Amir et al. \cite{amir2021deep}, which requires no training and offers reasonable results.
Note that our key contribution in attention masking is orthogonal to this clustering choice.

\textbf{Image-to-text generation with CLIP.}
The recent advent of CLIP leads to new approaches in text-image tasks, including generating text from an input image.
ClipCap \cite{mokady2021clipcap} trains a mapping network that joins CLIP with a pretrained language model, GPT-2, and performs image captioning with faster training.
ZeroCap \cite{tewel2022zerocap} performs zero-shot image caption by optimizing the value matrix $V$ in each attention module in GPT-2 to guide the embedding of the output text toward the target image's embedding. The output texts display knowledge learned from CLIP's vast and diverse training set, such as names of celebrities and pop culture references. This is a new ability unseen in older image captioning methods. Note that our contribution is not directly in text generation, rather we focus on inferring semantic segments and mapping them to CLIP's latent space.

\section{Approach}
\label{sec:method}
\begin{figure*}
\centering
  \includegraphics[scale=0.965]{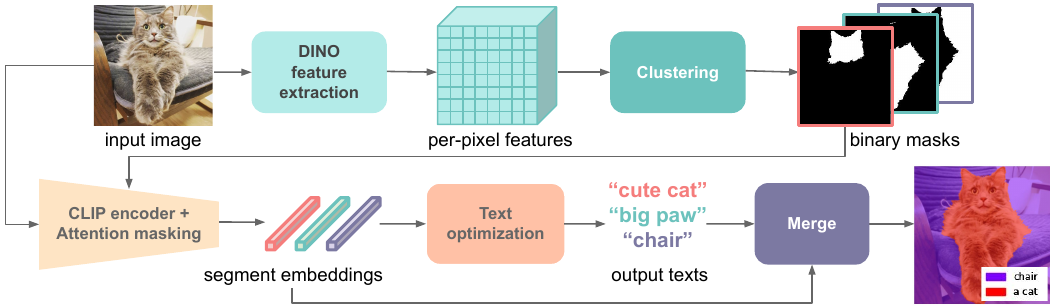}
  \vspace{-.2em}
  \caption{\textbf{Pipeline Overview.} Our method first segments an input image by clustering learned per-pixel features extracted from DINO. The input image is then fed into CLIP's image encoder. In this step, the produced segmentation masks are used to modify CLIP's attention to provide embeddings that are more focused to each segment. The resulting embeddings are then used to optimize a trained language model to generate texts closest to these embeddings. Lastly, segments with similar embeddings and text outputs are merged together to form more coherent segmentation results.}
  \label{fig:overview}
  \vspace{-1.em}
\end{figure*}

Given an input image, our goal is to partition this image into semantic segments and label each segment using words in natural language.
Our framework consists of four stages: 1) we identify segment candidates based on clustering deep per-pixel features of DINO-ViT \cite{caron2021emerging}, 2) we map each segment to an embedding in the CLIP's visual-language space using our proposed attention masking technique, 3) we translate each CLIP embedding into words by optimizing a generative language model with an existing technique, ZeroCap \cite{tewel2022zerocap}, and 4) we merge segments with similar semantics.

\subsection{Finding segment candidates with DINO} \label{sec:finding_segments}
The goal of this step is to partition the input image into small over-segments, which will be merged in the final step. To do so, we first extract spatial features of the input image from DINO-ViT. In particular, we use the ``key'' values from the last attention layer as the features (following \cite{amir2021deep}), which have a total dimension of (\#patch$\times C$). Unlike in standard use of ViT, we use a small stride of two instead of the patch size, resulting in a dense feature map ($\frac{H}{2}$$\times$$\frac{W}{2}$$\times$$C$).


Given this dense feature map, we initially assign each feature vector ($1\times$$C$) its own cluster and perform agglomerative clustering by repeatedly merging any two clusters with the smallest combined feature variance. We stop this process when the target number of clusters $n = 20$ is reached, and we additionally merge clusters with similar feature vectors based on their cosine similarity, detailed in Appendix \ref{apx:clustering}. The output segments from this step may break single objects into small parts, which lack semantic meanings by themselves. However, our decision to oversegment first allows merging in the semantic space of CLIP later on, which takes into account both vision and language semantics and can be done with simple thresholding.

\subsection{Transforming segment candidates into CLIP's vision-language embeddings}
\label{sec:region-emb}
To map a given segment to CLIP's vision-language space, our idea is to feed the entire input image into CLIP's image encoder while masking some of the encoder's attention layers with an alpha mask corresponding to the given segment. 
One major consideration is which layers should the masking be applied to properly balance global and local contexts. This turns out to be challenging: masking in earlier layers destroys global contexts, whereas masking in later layers eliminates local contexts. This finding agrees with several studies \cite{zhou2021deepvit, xue2022deeper, gong2021vision} showing that vision transformers trained for classification suffer from an ``attention collapse,'' where the attention in deeper layers becomes near uniform and all tokens converge to the same value. 
Another study \cite{shen2021much} also suggests that CLIP may lack the ability to maintain local information. In Appendix \ref{apx:clip-visualize}, we show how CLIP's attention maps become less localized in later layers.

Masking in the middle layers also performs poorly because different objects still require different degrees of context balancing depending on how salient they are in the scene. For example, the embedding of a small, obscure object in the background can be dominated by global contexts, which describe other prominent objects in the scene. As a result, these small objects may require more de-emphasis of the global contexts for their semantics to emerge.

Based on this observation, we propose a simple technique to estimate the saliency of each segment and use it to modulate how much global information should be removed or subtracted from individual tokens during attention masking. We next explain how we apply masking to the attention module, and then our \emph{global subtraction} technique.

\vspace{-.7em}
\subsubsection{Masking in self-attention module}\vspace{-.2em}
Given a logit vector $x \in \mathbb{R}^n$ and a flattened mask $M \in [0, 1]^n$, we first define the masked softmax operator as:
\begin{equation}
\mathrm{MaskedSoftmax}(x, M) = \frac{e^{x} \odot M}{\sum_{j=1}^{n} \left ( e^{x_j} \times M_j \right )},
\end{equation}
where $\odot$ denotes the element-wise multiplication.
To mask a standard attention module, we compute $A_i^\text{masked}=\mathrm{MaskedSoftmax}\left(Q_{i}K^{T} / \sqrt{d_{k}}, M\right)V$ for \emph{every} token $i$.
In practice, when the mask size is larger than the visual patch grid, we first downsample $M$ to the same size using area interpolation.
We also prepend one extra element to the flattened $M$ for the global token, which is always set to one in our algorithm, and thus $\text{\#tokens} =\text{\#patches} + 1$.

\begin{figure}
\centering
  \includegraphics[scale=0.82]{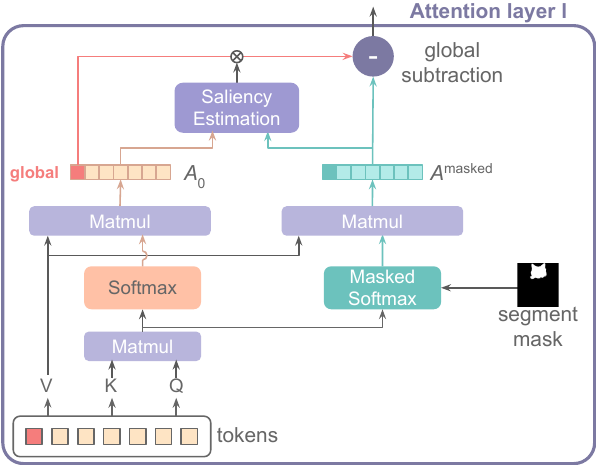}
  \caption{\textbf{Attention masking and global subtraction.} To encode a segment into CLIP's space, we pass the input image into CLIP's image encoder and mask self-attention map in some layers with the segment's mask. We apply this masking inside masked softmax function while still computing normal softmax. Cosine similarity between masked and unmasked output is used to estimate the saliency of the region. This similarity determines how much global context needs to be reduced in global subtraction.}
  \label{fig:attention_masking}
\end{figure}

\subsubsection{Global subtraction}
To balance global and local contexts in our output embedding, we design a proxy function that estimates the ``saliency'' of each segment or the segment's presence in the global contexts. This value will be used to determine how much global contexts should be removed from the attention output. Note that our saliency value is only defined with respect to the attention mechanism in CLIP and is unrelated to other uses of ``saliency'' in the literature \cite{4270292, 6112774}. 

We perform the following operations separately for each attention layer $l$ using its own saliency value $\mathcal{S}_l$. We compute $\mathcal{S}_l$ as the cosine similarity between the masked attention output $A^\text{masked}$ and unmasked attention output $A= \mathrm{Softmax}(QK^T/\sqrt{d_k})V$ at layer $l$, averaged over all tokens (we omit the subscript $l$ from $A$ and $A^\text{masked}$ for simplicity):
\begin{equation}
    \mathcal{S}_l = \frac{1}{\text{\#tokens}}\sum_{i = 1}^{\text{\#tokens}}\mathrm{cossim}(A_i, A_i^\text{masked}).
\end{equation}

The output of the attention layer $l$ is computed by subtracting the unmasked attention output of the global token $A_\text{0}$ from the masked attention $A^\text{masked}$:
\begin{align}
    A^\text{out} &= A^\text{masked} - w A_0,\\
    \label{eqn:salience_est}
    \text{where } w &= \exp(-(\mathcal{S}_l+1)^2 / 2\sigma^2).
\end{align}
This global subtraction weight $w$ is computed by applying a Gaussian function with a standard deviation $\sigma$ to $(\mathcal{S}_l+1)$, making $w$ highest when $\mathcal{S}_l=-1$. In other words, when an object is \emph{not} salient, 
we remove \emph{more} global contexts from its attention output. 
We apply these masking operations starting from attention layer 21, which is chosen empirically, to the last layer 24. Finally, the embedding of our segment is the output from our masked CLIP's encoder, which additionally applies linear projection to the global token value from the last attention layer.

\subsection{Text generation from CLIP's embedding}\label{sec:textgeneration}
To translate our segment embedding in CLIP's joint space into words, we use an existing image-to-text generation algorithm, ZeroCap \cite{tewel2022zerocap}.
This method uses a pretrained language model GPT-2 \cite{radford2019language} along with CLIP to optimize for a sentence that describes an input image. This is done by optimizing specific activations of GPT-2 (K and V matrices) to complete an initial prompt of  ``Image of a ...'' and minimizing the difference between the output sentence and the input image in CLIP's joint space.



\subsection{Merging segment candidates} \label{sec:merge}
In this step, we merge segments that are semantically similar or small segments that may not be so meaningful by themselves from our oversegmentation. 
We compute the similarity score between two segments using the average of two measures: 1. the cosine similarity between their visual embeddings (Section \ref{sec:region-emb}) and 2. the cosine similarity between their predicted texts' embeddings computed from CLIP's text encoder. 
In our implementation, we also reduce the number of merging combinations by limiting the pairing option. In particular, we first continue the agglomorative clustering in the first step (Section \ref{sec:finding_segments}) until there is a single cluster representing the entire image. By keeping track of the merging history, we obtain a binary tree where each node represents a segment and each parent is the merged segment of its children. We limit the pairing to only between siblings in this tree and recursively merge segments up the tree when their similarity score is at least $\tau_\text{merge}$. The final embedding of a merged segment is computed by passing the corresponding merged mask through the embedding pipeline (Section \ref{sec:region-emb}) and is then used to generate the final predicted text.



\section{New Evaluation Protocol}  \label{sec:eval}
This section introduces new metrics to evaluate the quality of the output segments and their corresponding text labels.
To overcome the evaluation challenges due to the use of synonyms or the difference in label granularity, such as ``car'' vs ``wheel,'' we first map the predicted labels to the predefined ones in the test set (Section \ref{label-re}) and verify the reassignment using thresholding or human evaluation (Section \ref{sim-thres}) before applying standard metrics such as IoU.


\subsection{Label reassignment} \label{label-re}
Given a predicted segment $S_i$ and its predicted text $T_i$, our goal is to relabel $S_i$ with $T^*_i$, which should be one of the test labels. We describe two reassignment techniques based on text-to-text and segment-to-text similarity.

\myparagraph{Text-to-text similarity (TT).} 
This technique relabels $S_i$ with the ground-truth label that is closest to $T_i$ in the embedding space of Sentence-BERT \cite{reimers-2019-sentence-bert}, a pre-trained text encoder widely used in NLP for computing text similarity \cite{gao2021simcse,li2020sentence,choi2021evaluation}.
Formally, the new label $T^*_i$ is computed by
\begin{align}
    T^*_i &= \argmax_{t \in T^\text{gt}} \left[ \mathrm{cossim}^\text{SBERT}(T_i, t)\right],
\end{align}


\myparagraph{Segment-to-text similarity (ST).}
This technique relabels $S_i$ with the ground-truth label that is closest to $S_i$ in the CLIP's joint image-text space \cite{hessel2021clipscore, kawar2022imagic}. That is, 
\begin{align}
    T^*_i &= \argmax_{t \in T^\text{gt}} \left[ \mathrm{cossim}^\text{CLIP}(S_i, t)\right],
\end{align}
where $\text{cossim}^\text{CLIP}(s, t)$ uses CLIP's image encoder for $s$ and text encoder for $t$. Note that this relabeling is commonly used in open-vocabulary settings \cite{liang2022open, ghiasi2021open}, but it does not consider our predicted label $T_i$ during relabeling. Nonetheless, this technique is still valuable as it offers a complementary assessment that does not involve text generation, which is based on prior work (Section \ref{sec:textgeneration}), or text-to-text mapping, which can be challenging and ambiguous even for human evaluators (Section \ref{sec:analysis}).


\subsection{Reassignment verification}\label{sim-thres}
For evaluation, we need to verify that the reassigned label $T^*_i$ is sufficiently close to the original label  $T_i$ or its segment $S_i$. We provide two kinds of verification. The first is based on simple thresholding on the cosine similarity using $\tau_\text{SBERT}$ and $\tau_\text{CLIP}$ for TT and ST reassignments, respectively (Appendix \ref{apx:thresholds}). The second involves human judgement, in which we ask human evaluators to rate how well the reassigned label describes its segment on a scale of 0-3, ranging from 0: incorrect, 1: partially correct, 2: correct but too general/specific, 3: correct. The full definitions are in Appendix \ref{apx:userstudy}. Multiple thresholds will be used to report scores.




\subsection{Metrics} \label{sec:metrics}
\vspace{-0.5em}
\myparagraph{Segmentation IoU} evaluates the quality of the output segments in terms of Intersection-over-Union (IoU) against the ground-truth segments in each test image. 
Given a set of predicted segments with reassigned labels $T^*$, segments with the same label $T^*$ are merged to form a single segment for the label. Then, IoU for each image can be computed  using a standard protocol \cite{everingham2010pascal}.

\myparagraph{Segment Recall}
measures how many objects labeled in the ground truth are discovered. This metric disregards any extra labels predicted by our method that are not part of the ground truth labels. We consider each merged segment of the same reassigned label a True Positive if its IoU against the corresponding ground-truth segment is greater than $\tau_\text{IoU}$.
Segment Recall is the rate of True Positive over the number of grounding segments. 



\myparagraph{Text Generation Quality} 
measures the quality of text generation given an oracle segmentation. 
That is, we feed each \emph{ground-truth} segment into our model and compute the cosine similarity between our predicted label and the ground-truth label.
If the value is higher than $\tau_\text{SBERT}$, it is considered a True Positive. The score is the True Positive rate over the entire test set. This metric evaluates our attention-masking and text generation components independent of the segment generation process (Section \ref{sec:finding_segments}). 

\begin{figure*}
\centering
 \includegraphics[scale=0.92]{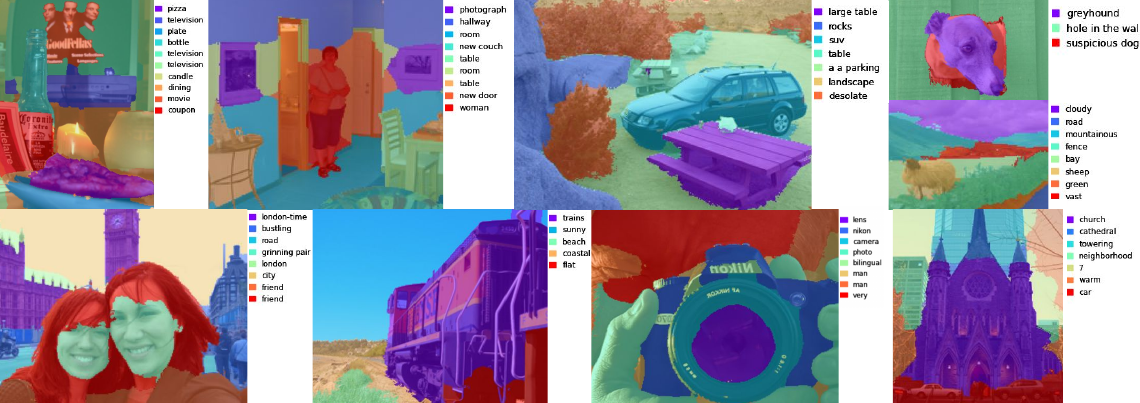}
  \caption{\textbf{Qualitative results.} Our method gives reasonable segments and outputs free-language text labels representing all regions. The labels can describe regions by different kinds of descriptions, such as object names, facial expressions, locations, car models, or even animal breeds.}
  \label{fig:results}
  \vspace{-1.5em}
\end{figure*}

\begin{figure*}
\centering
  \includegraphics[scale=1.04]{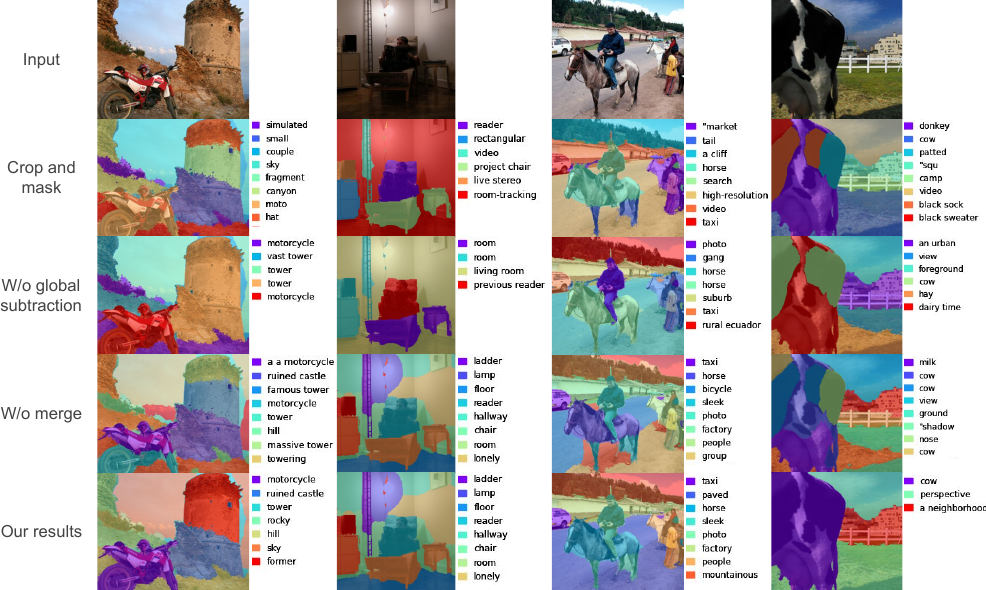}
  \caption{\textbf{Ablation results.} Comparison between the results from different segment encoding methods. 
  The crop-and-mask baseline often outputs text labels that is not relevant to segments/input images.
  Our method without global subtraction suffers from global leak and often mislabels non-salient objects.
  Without semantic merging, text outputs look good, but it tends to over-segment.
  }
  \label{fig:ablation}
\end{figure*}

\section{Experiments}\vspace{-.3em}

\vspace{-.3em}
\myparagraph{Datasets.}
We evaluate our results on two commonly used segmentation datasets: Pascal Context \cite{mottaghi2014role} and Pascal VOC 2012 \cite{everingham2010pascal}. 
Pascal Context contains 5,000 validation images with segmentation ground truths of 459 object classes for scene segmentation task. 
We use PC-59, the commonly used subset with 59 most common objects, following \cite{ghiasi2021open,xu2022groupvit,liang2022open}, as well as the full PC-459.
Pascal VOC (PAS-20) is a segmentation dataset with 1,500 image-segment validation pairs of 20 object classes.
For both datasets, we report our results on the validation splits, as the test splits are not publicly available, but the validation splits were never used for hypertuning.
For comparison with our own variations and a crop-and-mask baseline, we test on the first 1,000 images of Pascal Context dataset and full 1,500 image for Pascal VOC dataset (Section \ref{sec:ablation}). We use the full datasets when compared to prior work (Section \ref{sec:groupvit}). 



\subsection{Zero-guidance segmentation results} 
We present our qualitative results in Figure \ref{fig:teasor}, \ref{fig:results}.
Our method can discover semantic segments and densely label them with diverse types of labels, including names of objects, animal breeds, facial expressions, and places. More results are in Appendix \ref{apx:results}. 
In Table \ref{tab:quan results}, we report IoU and Recall scores using different reassignment and verification techniques (Section \ref{sim-thres}), computed on 1,000 randomly sampled images from PC-59. We observe that ST tends to perform reassignment better than TT, as evident by its higher scores. 
Reassigning words like `leg' to the correct animal class in the ground-truth set can be challenging when relying solely on text (TT), as it lacks any additional context. But ST can access other visual information within the segment, which better facilitates reassignment. 

\begin{table}[]
\caption{Quantitative results on 1,000 random images from PAS-59's validation split. We use constants ($\tau_\text{SBERT}$ and $\tau_\text{CLIP}$) and multiple human verification scores (h) for thresholding.}
\vspace{-0.5em}
\label{tab:quan results}
\resizebox{\columnwidth}{!}{
\setlength{\tabcolsep}{0pt}
\begin{tabular}{l@{\extracolsep{6pt}}ccc@{\extracolsep{6pt}}cccc}
\toprule
 & \multicolumn{3}{c}{\textbf{Text-text reassign.}}  & \multicolumn{4}{c}{\textbf{Segment-text reassign.}}                                \\ \cline{2-4} \cline{5-8}
 \\[-1em]
\textbf{Threshold:} && const. & h $ \geq 1$ & const. &  h $ = 3$ &  h $ \geq 2$ & h $ \geq 1$   \\  \midrule
IoU    && 11.2     & 11.0     & 19.3      & 14.2    & 20.9      & 22.7        \\
Recall    && 10.3     & 9.8     & 18.0      & 13.2    & 18.0      & 19.4        \\
 \bottomrule
\end{tabular}}
\end{table}

\begin{table}[]
\centering
\caption{Distribution of human rating scores on the quality of the predicted labels (0: incorrect, 1: partially correct, 2: correct but too general/specific, 3: correct).}
\vspace{-0.5em}
\label{tab:labelq}
\resizebox{\columnwidth}{!}{
\setlength{\tabcolsep}{12pt}
\begin{tabular}{ccccc}
\toprule
\textbf{Human rating} & 0 & 1 & 2 & 3  \\ \midrule
\% of labels    & 36.0 & 20.8 & 23.9  & 19.3 \\ \bottomrule
\end{tabular}}
\vspace{-1em}
\end{table}


\subsection{User study}
We evaluate the quality of predicted labels using human evaluation. 
Each segment and its predicted label were shown to three distinct human evaluators, who were asked to rate how well the label describes the segment on a scale of 0-3, similar to the process in Section \ref{sim-thres} except we show the predicted label $T_i$ instead of the reassigned label $T_i^*$. Full details and the score definitions are in Appendix \ref{apx:userstudy}.


Table \ref{tab:labelq} shows that human evaluators found about 43\% of our results to be `correct' or `correct but too generic/specific' and 64\% to be at least `partially correct.' 

\begin{figure}
\centering
 \includegraphics[scale=0.77]{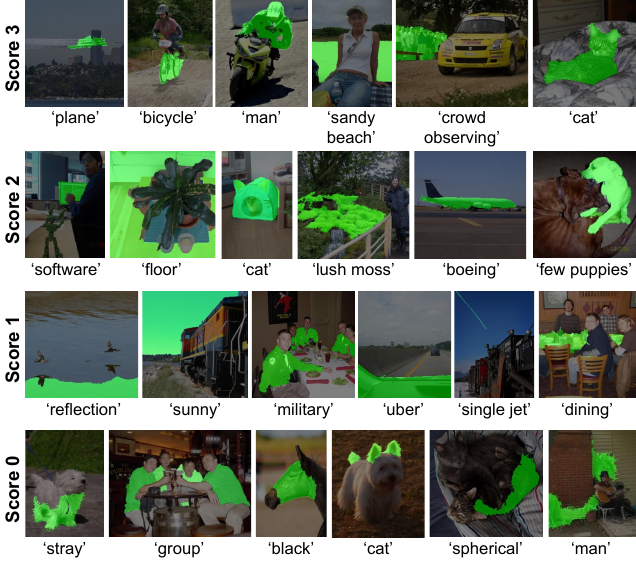}
  \vspace{-1.8em}
  \caption{Examples of segmentation results that were evaluated by human evaluators with scores ranging from 3 (correct) to 0 (incorrect). The definitions of the scores can be found in Appendix \ref{apx:userstudy}.} 
  \vspace{-.5em}
  \label{fig:user_study_adj_failure}
\end{figure}

We provide example images and their scores given by the human evaluators in Figure \ref{fig:user_study_adj_failure}. According to the result, most of our score-0 labels are single-word adjectives, such as `black', or collective nouns, such as `group'. Another kind of score-0 labels is caused by biases toward stereotypical appearances of objects, such as when a pet dog was mislabeled as `stray' due to its shabby appearance (row 4). Some of score-1 labels correspond to descriptions or abstract nouns that are related to their segments but may not fully describe them, such as `reflection', `dining', and `sunny', and some other labels describe specific but incorrect types of objects, such as `uber' or `military'. Most of our score-2 labels are nearly accurate, but the segments may incompletely or excessively cover the referred objects, such as `lush moss' and `few puppies' (row 2). Most of our score-3 labels accurately represent their segments, such as `plane', and they can be descriptive even on background objects, such as `sandy beach' and `crowd observing', unlike labels from traditional segmentation methods.

\subsection{Ablation study} \label{sec:ablation} 

We compare our method with alternative attention-masking methods, which include 1) cropping and masking the input image to fit the segment region \cite{xu2021simple}, 2) our method without global subtraction, and 3) our method without the merging step.
Note that we also apply the same merging in the crop-and-mask baseline for a fair comparison. We show the results in Figure \ref{fig:ablation} and in Table \ref{tab:ablation}.

In Figure \ref{fig:ablation}, the crop-and-mask baseline often returns text labels that are unrelated to the segments, like `video' in column 2-4.  
Without global subtraction, our method often fails to recognize objects in the background due to the leak of global contexts. For example, the sky in column 1 is labeled `tower', and almost everything in column 2 is labeled `room'.
Our full pipeline yields reasonable results, and can label `ladder', `lamp', `floor', and `reader' correctly.

In Table \ref{tab:ablation}, we report results based on ST reassignment and constant thresholding. Our method outperforms all alternative masking techniques in terms of IoU, Recall, and Text Generation Quality scores on PC-59. Global subtraction also helps improve both $\text{IoU}_\text{CLIP}$ by 3.0-3.1 points. On PAS-20, our method achieves a slightly lower IoU than not using global subtraction (1.1 lower). Upon inspection, we observe that the 20-class PAS-20 tends to label only a few foreground objects while ignoring much of the background (see Appendix \ref{apx:voc}), and not using global subtraction may preserve the embedding of these few objects better. This bias toward primary objects, however, would not be beneficial if the goal is to discover \emph{all} semantic objects.

\begin{table}[]
\caption{Ablation study of our CLIP's mask attention technique. IoU and Recall are computed with ST and constant thresholding.
}
\vspace{-0.5em}
\label{tab:ablation}
\resizebox{\columnwidth}{!}{
\setlength{\tabcolsep}{4pt}
\begin{tabular}{lccc@{\extracolsep{6pt}}c@{\extracolsep{6pt}}c}
\toprule
 & \multicolumn{3}{c}{\textbf{PC-59}}  & \textbf{PC-459}& \textbf{PAS-20}                         \\ \cline{2-4} \cline{5-5} \cline{6-6}\\[-1em]
\multicolumn{1}{c}{Method} & \rule{0pt}{2ex}$\text{IoU$_\text{c}$}$ & $\text{Recall$_\text{c}$}$ &  TGQ  & $\text{IoU$_\text{c}$}$ &  $\text{IoU$_\text{c}$}$  \\ \midrule
Crop and Mask               & 12.1       & 10.2 & 16.9 & 5.4 & 14.0        \\
Ours w/o glob sub.        & 14.5         & 15.0 & 11.8 & 7.2 & 21.2          \\
Ours w/o merge              & 16.4       & 11.8 & - & 10.2  & 18.3         \\
Ours                        & 17.5       & 15.0 & 19.0 & 11.3  & 20.1       \\ \bottomrule
\end{tabular}}
\end{table}

\subsection{Comparison to zero-shot open-vocab baseline} \label{sec:groupvit}
As a reference, we provide a comparison with GroupVit \cite{xu2022groupvit}, which solves a related but different segmentation problem, \emph{open-vocabulary segmentation}. 
This task requires text queries to specify which objects to segment, although the queries can be arbitrary or unseen during training. 
Our method, on the contrary, predicts arbitrary text labels at inference time and is not directly comparable using the same standard benchmarks. Nonetheless, our proposed relabeling procedure can allow useful comparative analysis against open-vocabulary baselines on the same benchmarks. 

\begin{figure*}
  \includegraphics[scale=1.04]{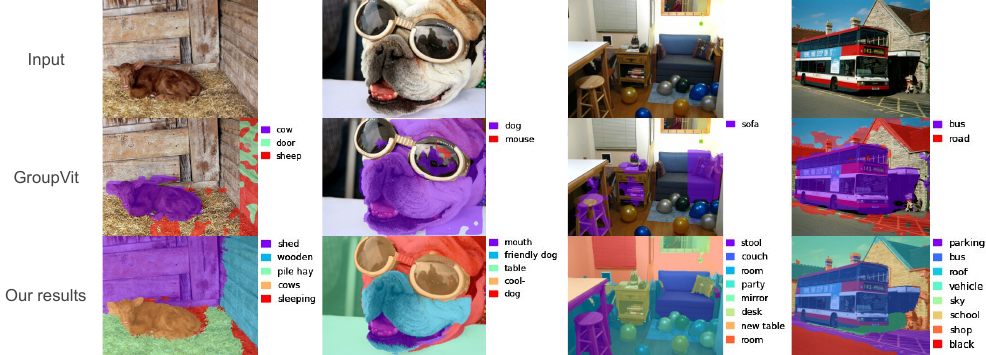}
\vspace{-.5em}
  \caption{\textbf{Qualitative comparison to GroupVit \cite{xu2022groupvit}.} 
  Despite achieving lower IoU scores, our method can discover objects beyond the labels in the dataset, such as `hay' and `mirror', and can provide more fine-grained labels, such as `stool'.
  }
  \label{fig:groupvit}
  \vspace{-.5em}
\end{figure*}

Table \ref{tab:groupvit} shows that GroupVit obtains a better IoU on PAS-20 with 20 classes. However, our method is significantly narrowing the gap on PC-59, especially with human-threshold IoU, and our IoU with constant-threshold even surpasses GroupVit's on challenging PC-459, which has much more classes (459).
Figure \ref{fig:groupvit} shows that our method can discover more objects and provide more fine-grained labeling, while GroupVit labels only a few objects and does not label every part of the image.


\begin{table}[]
\centering
\caption{Comparison to GroupVit \cite{xu2022groupvit}, which solves a related but different segmentation problem and requires input text queries. *denotes scores computed on 1,000 random test images. IoU$_\text{c}$ and IoU$_\text{h}$ are IoU with constant thresholding or human verification. GroupVit's numbers have been updated according to \cite{xu2023open}}
\vspace{-0.5em}
\label{tab:groupvit}
\resizebox{\columnwidth}{!}{
\setlength{\tabcolsep}{5pt}
\begin{tabular}{lcc@{\extracolsep{6pt}}ccc}
\toprule
                            & \multicolumn{3}{c}{\textbf{PC-59}}                               & \multicolumn{1}{c}{\textbf{PC-459}}  &   \multicolumn{1}{c}{\textbf{PAS-20}}   \\ \cline{2-4} \cline{5-5} \cline{6-6}\\[-1em]
 \multicolumn{1}{c}{Method} &  IoU$_\text{c}$  & IoU$_{\text{h } \geq 2}$  & IoU$_{\text{h } \geq 1}$ & $\text{IoU}_\text{c}$ & $\text{IoU}_\text{c}$  \\ \midrule
GroupVit \cite{xu2022groupvit} & 25.9      & -     & -      & 4.9     & 50.7 \\
Ours                           & 19.6      & 20.9*      & 22.7*     & 11.3   & 20.1  \\ \bottomrule
\end{tabular}}
\vspace{-1em}
\end{table}


\subsection{Comparison to supervised baselines}
\label{apx:baselines}
Table \ref{tab:baselines} presents an IoU comparison with existing supervised open-vocabulary baselines on three datasets based on the numbers presented in \cite{liang2022open}.
Unlike our approach, these methods require segmentation annotations (or pretrained segmentation models) during training and text queries to guide the segmentation. 
Our IoU scores in Table \ref{tab:baselines} are computed using segment-to-text IoU with a constant threshold $\tau_\text{CLIP} = 0.1$ and human score $\geq 1$.
There is still a gap in performance between our unsupervised method and these supervised baselines, though our method performs only slightly worse on the more challenging PC-459.

\begin{table}[]
\centering
\caption{IOU scores comparison between supervised open-vocabulary segmentation baselines (trained with segmentation labels) and our unsupervised method.}
\label{tab:baselines}
\vspace{-0.5em}
\resizebox{\columnwidth}{!}{
\setlength{\tabcolsep}{12pt}
\begin{tabular}{lccc}
\toprule
Method   & PAS-20 & PC-59  & PC-459 \\ \midrule
Lseg \cite{li2022language}          & 47.4   & -     &  -     \\
SimBaseline \cite{xu2021simple}     & 74.5   & -     &  -   \\
ZegFormer \cite{ding2022decoupling} & 80.7   & -     &  -     \\
OpenSeg \cite{ghiasi2021open}       & -      & 42.1  &  9.0   \\
OVSeg \cite{liang2022open}          & 94.5   & 55.7  &  12.4   \\  \midrule
Ours - $\text{IoU}_\text{c}$       & 20.1  &   19.6  & 11.3 \\ 
Ours - $\text{IoU}_{\text{h } \geq 1}$  & - &   22.7  & - \\ \bottomrule
\end{tabular}}
\end{table}
\begin{figure}
\centering
  \includegraphics[scale=0.615]{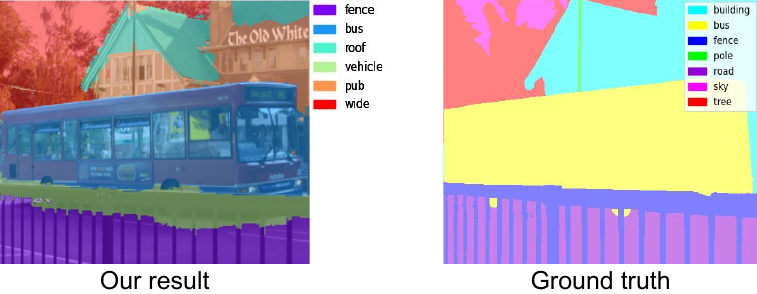}
  \vspace{-0.7em}
  \caption{
  \textbf{Label reassignment issue.} Our predicted labels `roof' and `pub' are correct but are not matched to the ground-truth class `building' during label reassignment.}
  \label{fig:result_analysis}
\end{figure}

\section{Discussion and Analysis} \label{sec:analysis}

\begin{figure}
\centering
    \includegraphics[scale=0.9]{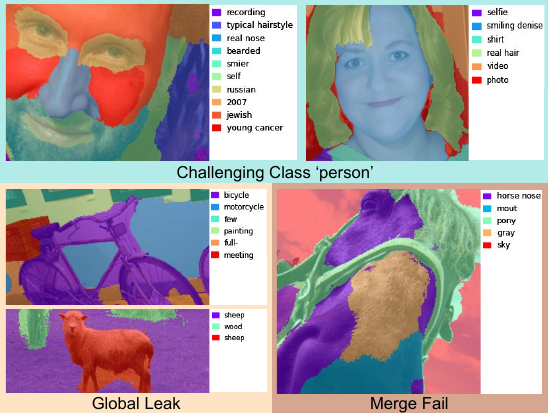}
  \vspace{-1.5em}
  \caption{\textbf{Failure cases.} 
  1) Classes that have many visually distinct parts, such as `person', are difficult to reassign labels correctly. 
  2) Background regions that share boundaries with salient objects are still prone to \emph{global context leakage}. 
  3) Semantic merging may fail when the text outputs of the same object give different descriptions.
  }
  \label{fig:failure_case}
  \vspace{-1.em}
\end{figure}

\myparagraph{Mismatched text labels during evaluation.} Evaluation in our new setup is still challenging, despite using label reassignment.
For example, in Figure \ref{fig:result_analysis}, our algorithm breaks down the `building' ground-truth segment into `roof' and `pub', which are correct.
But ST reassignment assigns `pub' to `sign', which is still technically correct but not counted toward our IoU score for `building'. 
Another problematic class is `person' whose parts like `face', `hair', `shirt' appear distinct in CLIP's space and may not be mapped to `person' (Figure \ref{fig:failure_case}).
To overcome this challenge, we may need a new kind of embedding space that understands the hierarchical nature of object parts.

\myparagraph{Global context leakage.} Some background segments that share boundaries with primary objects can be mislabeled due to the influence of global contexts as shown in Figure \ref{fig:failure_case}. 
Another problem that can cause context leakage is the low-resolution 24x24 image grid of CLIP visual tokens. As we downsample our segment masks to fit this grid, we lose masking precision and information can leak between neighboring segments.


\myparagraph{Merge fail due to different labels of the same object.}
Our over-segment outputs may use a wide variety of descriptions for the same object, such as car model and car color.
These segments may fail to merge into a single object during the merging step (see Figure \ref{fig:failure_case}).

\section{Conclusion} We have presented the first framework for \emph{zero-guidance segmentation}, a novel problem that seeks to segment and label an image using natural language automatically. We leverage two generalist models, DINO and CLIP, and propose a technique to map a given segment to CLIP's joint space by masking CLIP's attention, allowing zero-shot segmentation without the need for any segmentation dataset or fine-tuning. We also introduce a new evaluation protocol for this problem and will release our code.




\section*{Acknowledgements} 
This research was supported by PTT public company limited and SCB public company limited.

{\small
\bibliographystyle{ieee_fullname}
\bibliography{content/egbib}
}

\clearpage 
\appendix
\section*{Appendix: Zero-guidance Segmentation Using Zero Segment Labels}
\vspace{10pt}
In this Appendix, we provide additional details and experiments:
\begin{itemize}
    \item Section~\ref{apx:clip-visualize}: CLIP's self-attention visualization
    \item Section~\ref{apx:clustering}: Implementation details of our segment candidate finding method
    \item Section~\ref{apx:thresholds}: Thresholds used in metrics
    \item Section~\ref{apx:voc}: Limitations of Pascal VOC dataset for evaluation.
    \item Section~\ref{apx:hypertune}: Details on hyperparameters tuning
    \item Section~\ref{apx:userstudy}: User study
    \item Section~\ref{apx:results}: Additional results
    \item Section~\ref{apx:negimpact}: Potential negative societal impacts
\end{itemize}

\section{CLIP's Self-attention Visualization} \label{apx:clip-visualize}
\begin{figure}
\centering
  \includegraphics[scale=0.45]{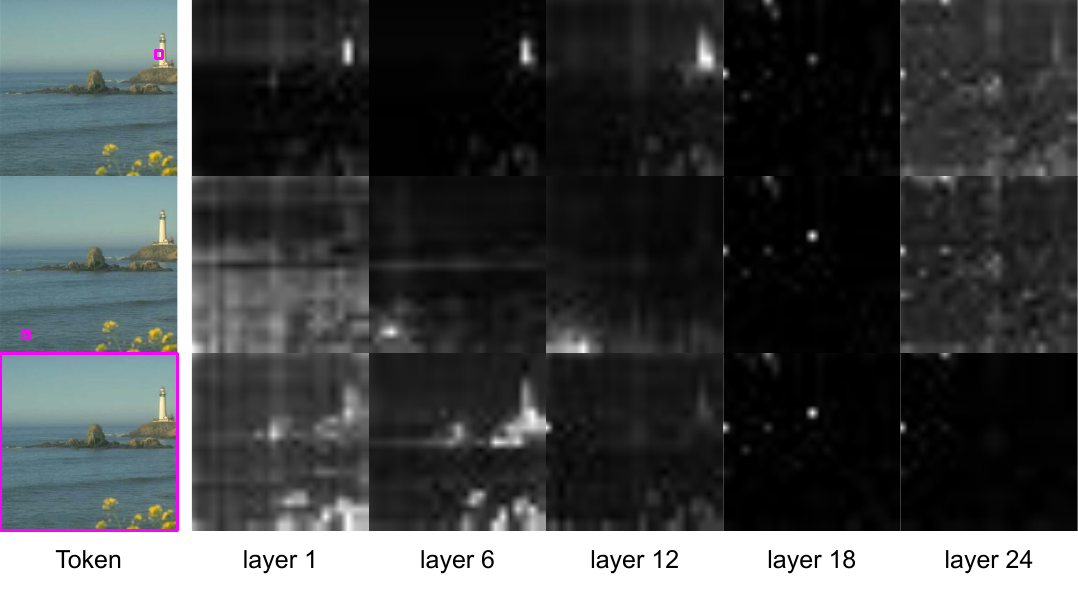}
  \vspace{-.5em}
  \caption{\textbf{Visualization of self-attention in CLIP's image encoder.} Each row shows the attention of the token of the pink patch across layers. The last row shows global token's attention.}
  \label{fig:self_attn}
  \vspace{-1.2em}
\end{figure}

Figure \ref{fig:self_attn} visualizes the self-attention maps of CLIP's image encoder across different layers. The self-attention maps appear to be meaningful in the earlier layers, i.e., the patch tokens mostly attend to regions that contain semantically similar pixels, and the global token attends to regions with prominent objects. However, the self-attention map appear more random and uninterpretable in the later layers.

\section{Finding Segment Candidates with DINO: Implementation Details} \label{apx:clustering}
We provide more implementation details for Section 3.1. 
We adopt DINO feature extraction method from Amir et al. \cite{amir2021deep}. The method first feeds an input image into DINO and extracts ``key'' values from the last attention layer as dense spatial features. 

After extracting the features, we partition the image into segments by clustering DINO's features.
We perform bottom-up clustering starting from each feature vector. The merging is done recursively by combining two clusters with the least combined variance. After this initial clustering, we end up with a binary tree where the root is the cluster of all the feature vectors. This binary tree structure is used as a heuristic to perform divisive clustering. Each node in the tree is represented by the average feature of its members. We prune the siblings whose cosine similarity score is over  $T_{Dino} = 0.9$. This yields a segmentation map with all leaf nodes of the binary tree as segments. The two-stage clustering algorithm is chosen to lessen the computation requirement since we start from a large number of spatial features ($111\times111$).

Following Amir et al, the segmentation map is then upsampled to input resolution and refined using DenseCRF as described in \cite{krahenbuhl2011efficient}. The Unary Energy is set as the normalized distance of each feature vector to all $k$ centroids, and the pairwise connection is fully-connected. 
Pairwise edge potentials are Gaussian kernels with location (pixel coordinates) as feature and Bilateral kernels with location and RGB values as features. 
Our implementation can be founded in the provided source code.

\section{Thresholds Used in Metrics}
\label{apx:thresholds}
\begin{figure}
\centering
  \includegraphics[scale=0.47]{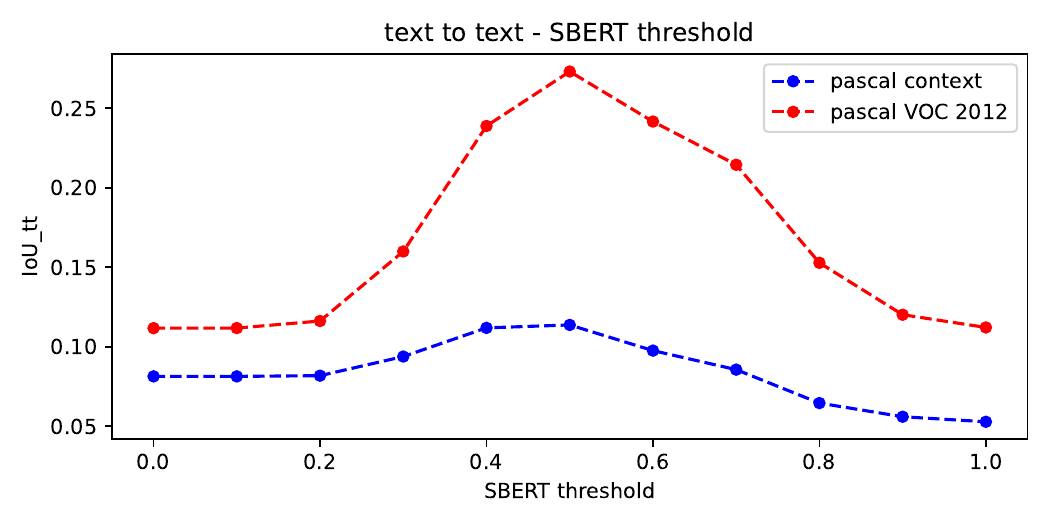}
  \caption{\textbf{Text to text IoU and SBERT threshold }}
  \label{fig:sbert_thres}
\end{figure}

\textbf{S-BERT text-to-text similarity threshold ($\tau_\text{SBERT}$).} 
\label{apx:text_sim}
We provide Text-to-text IoU (IoU$_\text{tt}$) scores with several $\tau_\text{SBERT}$ threshold values in Figure \ref{fig:sbert_thres} and Table \ref{tab:thres_sbert}. In the main experiment, when referred to a constant threshold, we select $\tau_\text{SBERT}=0.5$ as it represents an approximate minimum threshold that human evaluators use to determine if two sentences share a common topic, based on a user study \cite{cer2017semantic, gao2021simcse}. 



\begin{figure}
\centering
  \includegraphics[scale=0.47]{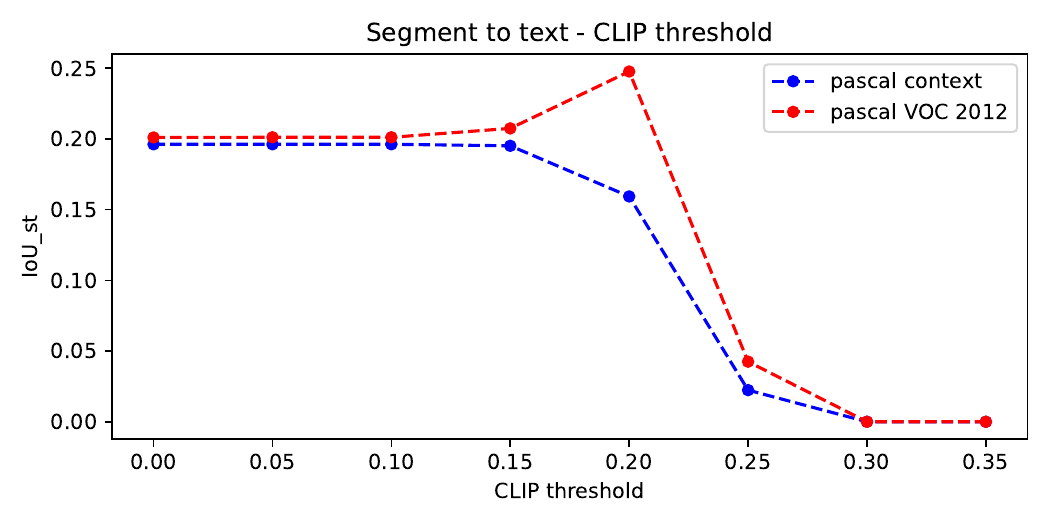}
  \caption{\textbf{Segment to text IoU and CLIP threshold }}
  \label{fig:clip_thres}
\end{figure}

\begin{figure}
\centering
 \includegraphics[scale=0.33]{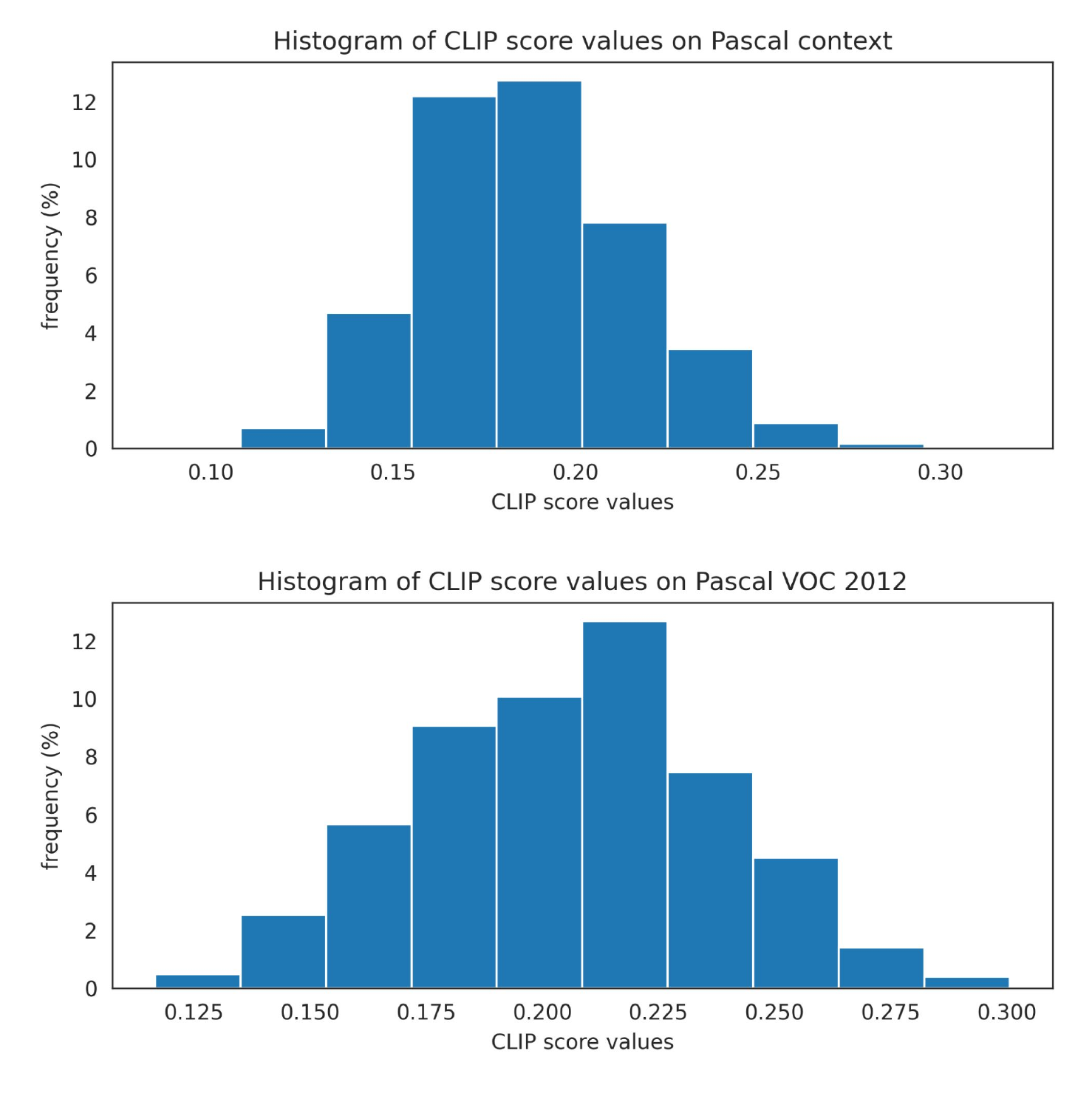}
  \caption{\textbf{CLIP similarity score distribution between the ground truth segment and the ground truth label}}
  \label{fig:clip_hist}
\end{figure}

\textbf{CLIP segment-to-text similarity threshold ($\tau_\text{CLIP}$).} 
We provide Segment-to-text IoU (IoU$_\text{st}$) scores with several $\tau_\text{CLIP}$ threshold values in Figure \ref{fig:clip_thres} and Table \ref{tab:thres_clip}.
Selecting the threshold $\tau_\text{CLIP}$ is more challenging, since there is no established consensus or user studies to rely on. Figure \ref{fig:clip_hist} shows histograms of CLIP similarity scores between ground-truth image segments and their corresponding ground-truth labels in Pascal Context and Pascal VOC datasets. Given the distributions, we select $\tau_\text{CLIP} = 0.1$ to be on the safe side to report Segment-to-text IoU scores in the main experiment. 

It is important to note that for our zero-guidance segmentation problem, the thresholds $\tau_\text{CLIP}$ and $\tau_\text{SBERT}$ are used in the label reassignment verification process (Section 4.2), which is part of the evaluation not the segmentation algorithm itself.
For a given algorithm, varying the threshold values can result in distinct performance profiles, e.g., a precision-recall curve, and several thresholds may be used together for the purposes of evaluation and comparison, as is common practice in the object detection literature \cite{zou2019object}.

\textbf{IoU threshold ($\tau_\text{IoU}$).}   \label{sec:iou_thres}
We use $\tau_\text{IoU} = 0.5$, which is commonly used in object detection tasks to determine if a predicted bounding box is `correct' compared to the ground truth \cite{zou2019object}.


\begin{figure}
\centering
  \includegraphics[scale=0.47]{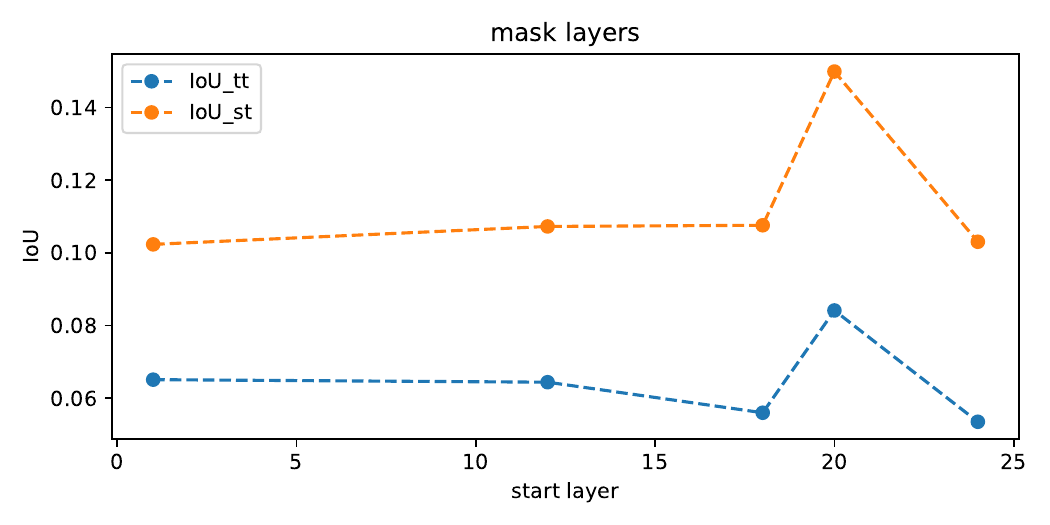}
  \caption{\textbf{Start masking layer selection}}
  \label{fig:start_ly}
\end{figure}

\begin{figure}
\centering
  \includegraphics[scale=0.47]{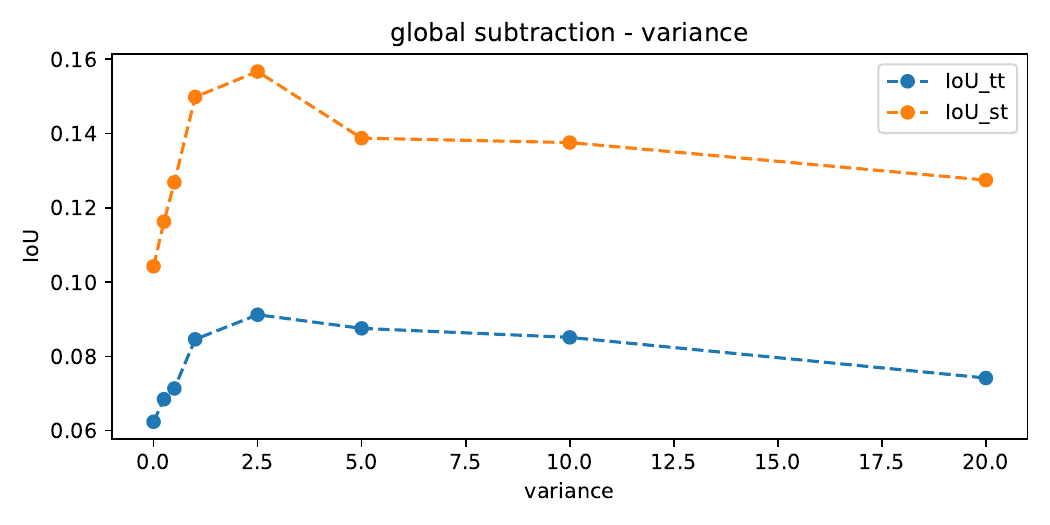}
  \caption{\textbf{Global subtraction's variance selection}}
  \label{fig:gs_sd}
\end{figure}

\begin{figure}
\centering
  \includegraphics[scale=0.47]{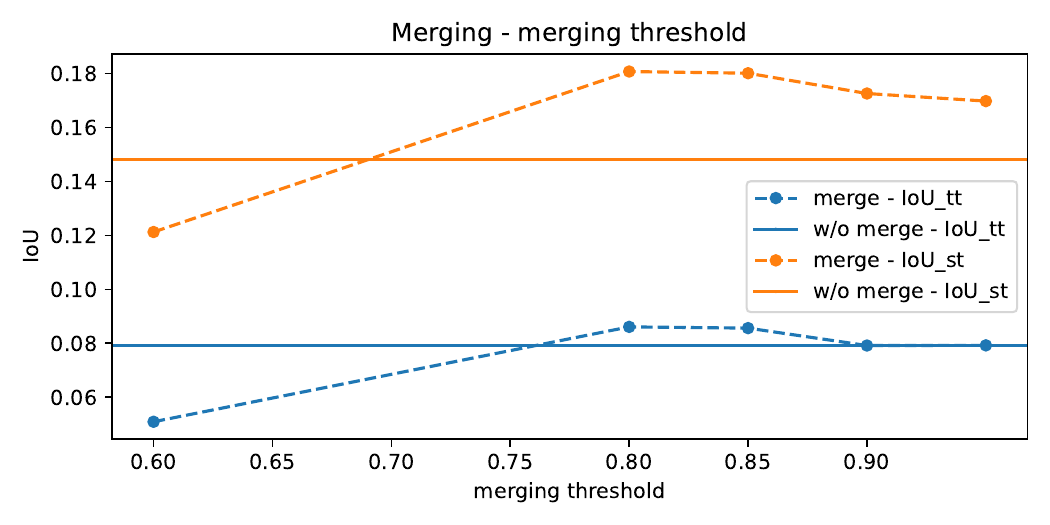}
  \caption{\textbf{Merging threshold selection }}
  \label{fig:merge_design}
\end{figure}





\begin{figure}
    \centering
    \includegraphics[scale=0.38]{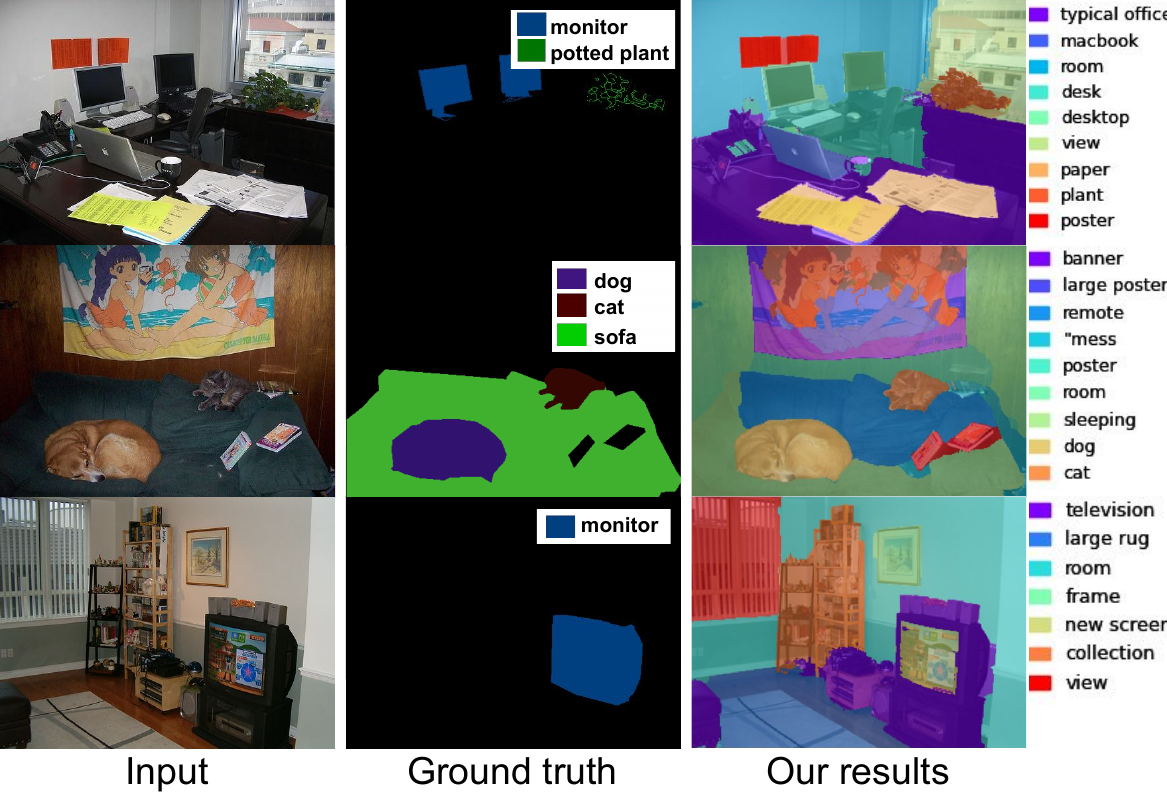}
\vspace{-.5em}
  \caption{An example of Pascal VOC segmentation dataset}
  \label{fig:2012}
\end{figure}


\section{Limitations of Pascal VOC for evaluating zero-guidance segmentation.} \label{apx:voc}
Evaluating zero-guidance segmentation performance using Pascal VOC (PAS-20) may not be ideal because PAS-20 has a very small number of labeled classes. In this dataset, many objects or sometimes the vast majority of regions in the images are left unlabeled as shown in Figure \ref{fig:2012}.
Our method can discover various objects not presented in the ground truth labels, such as `paper', `macbook', and `poster', but these are never counted towards any IoU scores.

\section{Hyperparameters Tuning} 
\label{apx:hypertune}

We present how our hyperparameters, which are the layer to start attention-masking, the global subtraction variance, and the merging threshold, are tuned.
Our tuning metrics are the Text-to-text IoU (IoU$_\text{tt}$) and Segment-to-text IoU (IoU$_\text{st}$) with the constant thresholds. The data used in this process are 100 randomly selected images from the Pascal Context's training split, which is never used for evaluation. Note here that there is no training involved in our pipeline.

The first parameter is the layer where attention masking starts. 
We found that masking from early layers erases all global context, resulting in poor results as context can be crucial for recognizing objects. Masking only the last layer also has poor results due to global leak. We found that masking attention of the last four layers (21-24) gives the best scores (see Figure \ref{fig:start_ly}).

Another important hyperparameter is the variance ($\sigma^2$) in the saliency estimation, which is used to determine the degree of global context subtracted from a region (see Equation 4). The higher the variance, the more global context is reduced. As seen in Figure \ref{fig:gs_sd}, the optimal spot is at $\sigma^2 = 2.5$. 

The last parameter is the merging threshold $\tau_{merge}$ used to decide which segment candidates to merge (Section 3.4). We found that $\tau_{merge} = 0.8$ returns the best scores on both IoU$_\text{tt}$ and IoU$_\text{st}$ on the tuning set (see Figure \ref{fig:merge_design}).

\begin{table}[]
\caption{Text-to-text IoU with several SBERT thresholds}
\vspace{-0.5em}
\label{tab:thres_sbert}
\resizebox{\columnwidth}{!}{
\setlength{\tabcolsep}{1.5pt}
\begin{tabular}{lccccccccrrr}
\toprule
               & \multicolumn{11}{c}{IoU$_\text{tt}$}                                       \\ \midrule
$\tau_\text{SBERT}$ & 0.0 & 0.1 & 0.2 & 0.3 & 0.4 & 0.5 & 0.6 & 0.7 & 0.8  & 0.9  & 1.0    \\ \midrule
Pascal Context & 8.1  & 8.1  & 8.2  & 9.4  & 11.2 & 11.4 & 9.8  & 8.6  & 6.5  & 5.6  & 5.3  \\
Pascal VOC     & 11.2 & 11.2 & 11.6 & 16.0 & 23.9 & 27.3 & 24.2 & 21.5 & 15.3 & 12.0 & 11.2 \\ \bottomrule
\end{tabular}}
\end{table}

\begin{table}[]
\caption{Segment-to-text IoU with several CLIP thresholds}
\vspace{-0.5em}
\label{tab:thres_clip}
\resizebox{\columnwidth}{!}{
\begin{tabular}{lcccccccc}
\toprule
                              & \multicolumn{8}{c}{IoU$_\text{st}$}                   \\ \midrule
$\tau_\text{CLIP}$  & 0.00 & 0.05 & 0.10 & 0.15 & 0.20 & 0.25 & 0.30 & 0.35 \\ \midrule
Pascal Context & 19.6 & 19.6 & 19.6 & 19.5 & 16.0 & 2.2  & 0.0  & 0.0  \\
Pascal VOC     & 20.1 & 20.1 & 20.1 & 20.8 & 24.8 & 4.2  & 0.0  & 0.0  \\ \bottomrule
\end{tabular}}
\end{table}

\section{User Study Implementation Details}
\label{apx:userstudy}
\begin{figure*}
\centering
 \includegraphics[scale=0.4]{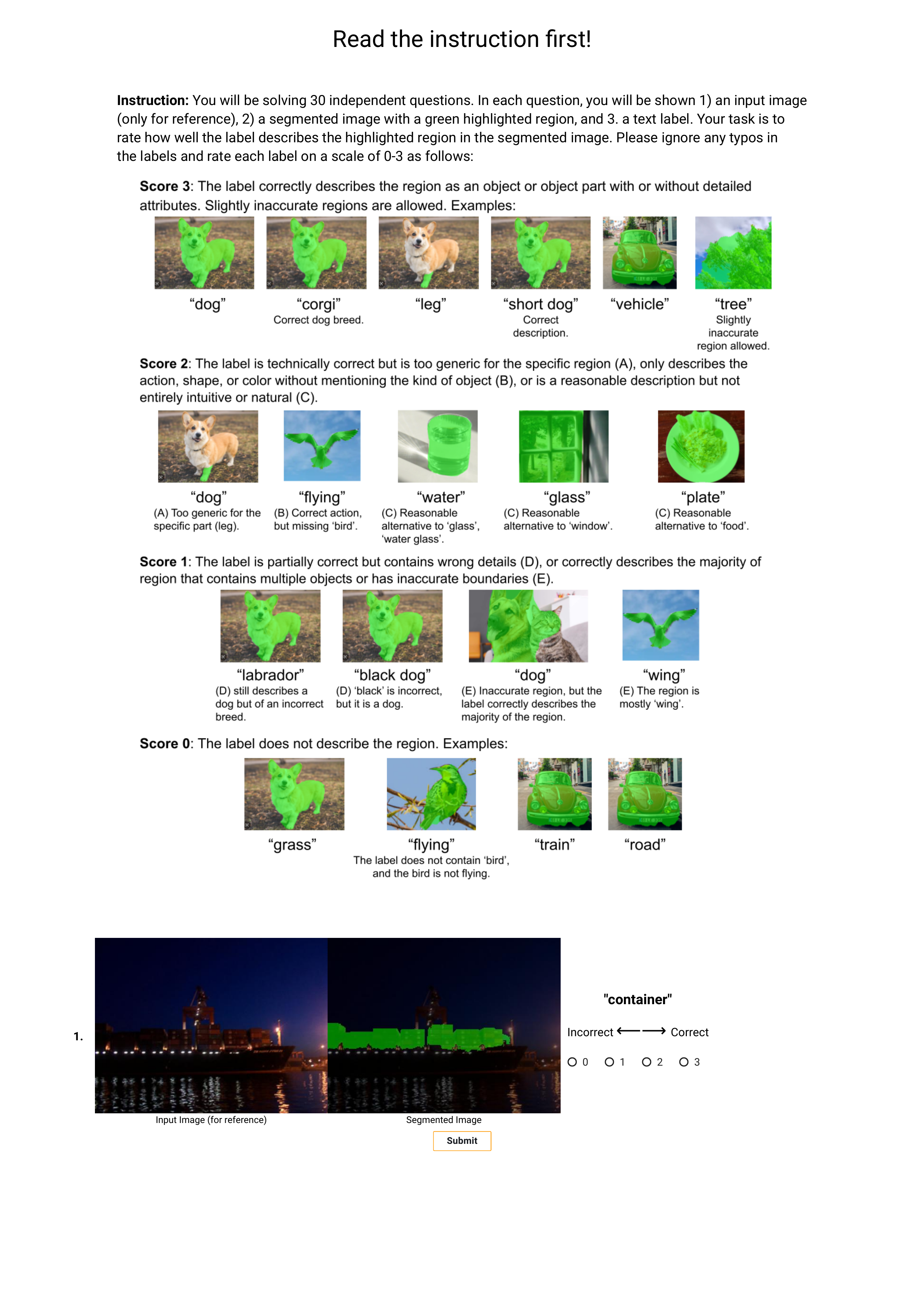}
  \vspace{-7.em}
  \caption{User interface for the user study with a full instruction, definitions, and examples of each score.} 
  \label{fig:user_study_ui}
\end{figure*}

We conducted a user study using Amazon Mechanical Turk. 
Each evaluation task contains a detailed instruction and 30 questions. We did not limit the number of tasks per human evaluator.
For each question, the evaluators were shown a predicted segment, in the form of a highlighted region, overlaid on an input image, and its predicted text label. The evaluators were then asked to rate how well the label describes the segment on a scale of 0-3, defined in the provided instruction as shown in Figure \ref{fig:user_study_ui}. 
There were a total of 23,076 questions, each evaluated by three different evaluators. The total number of unique evaluators was 429, and the average number of questions answered by the evaluators was about 155.
We calculated the scores (Section 4.3) for each of the three batches separately then reported the average.
We include the full instruction and a task example in Figure \ref{fig:user_study_ui}.

\section{Additional Results}
\label{apx:results}
\begin{figure*}
\centering
 \includegraphics[scale=0.85]{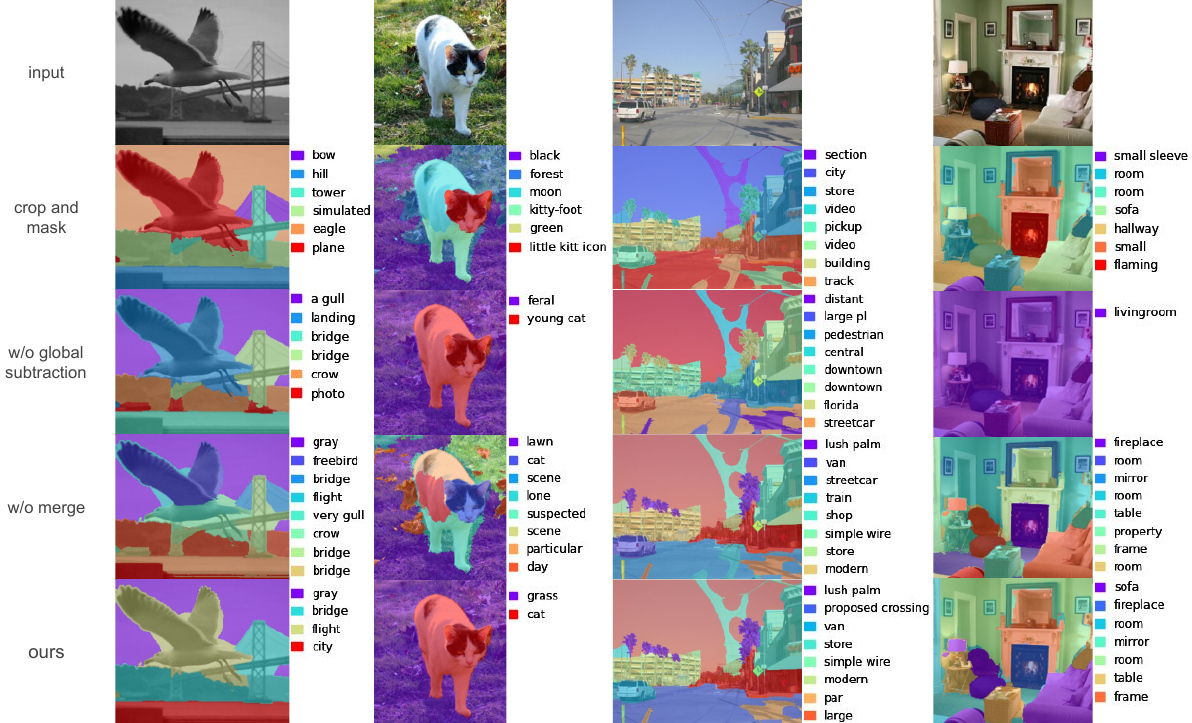}
  \caption{Qualitative ablation analysis} 
  \label{fig:ablation_apx}
\end{figure*}

\begin{figure*}
\centering
 \includegraphics[scale=0.82]{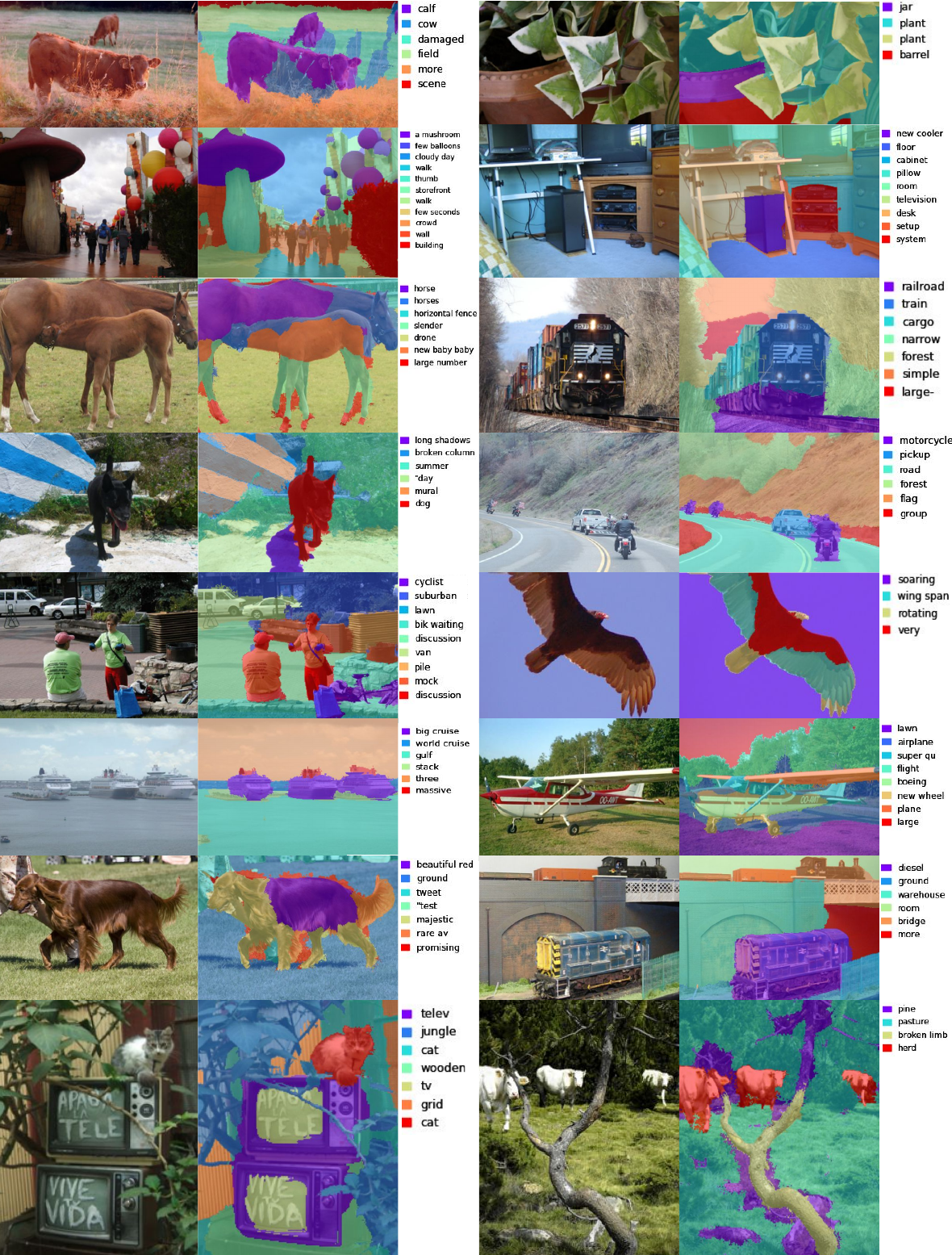}
  \vspace{.5em}
  \caption{Randomly sampled results from Pascal Context} 
  \label{fig:ran_context}
\end{figure*}

\begin{figure*}
\centering
 \includegraphics[scale=0.82]{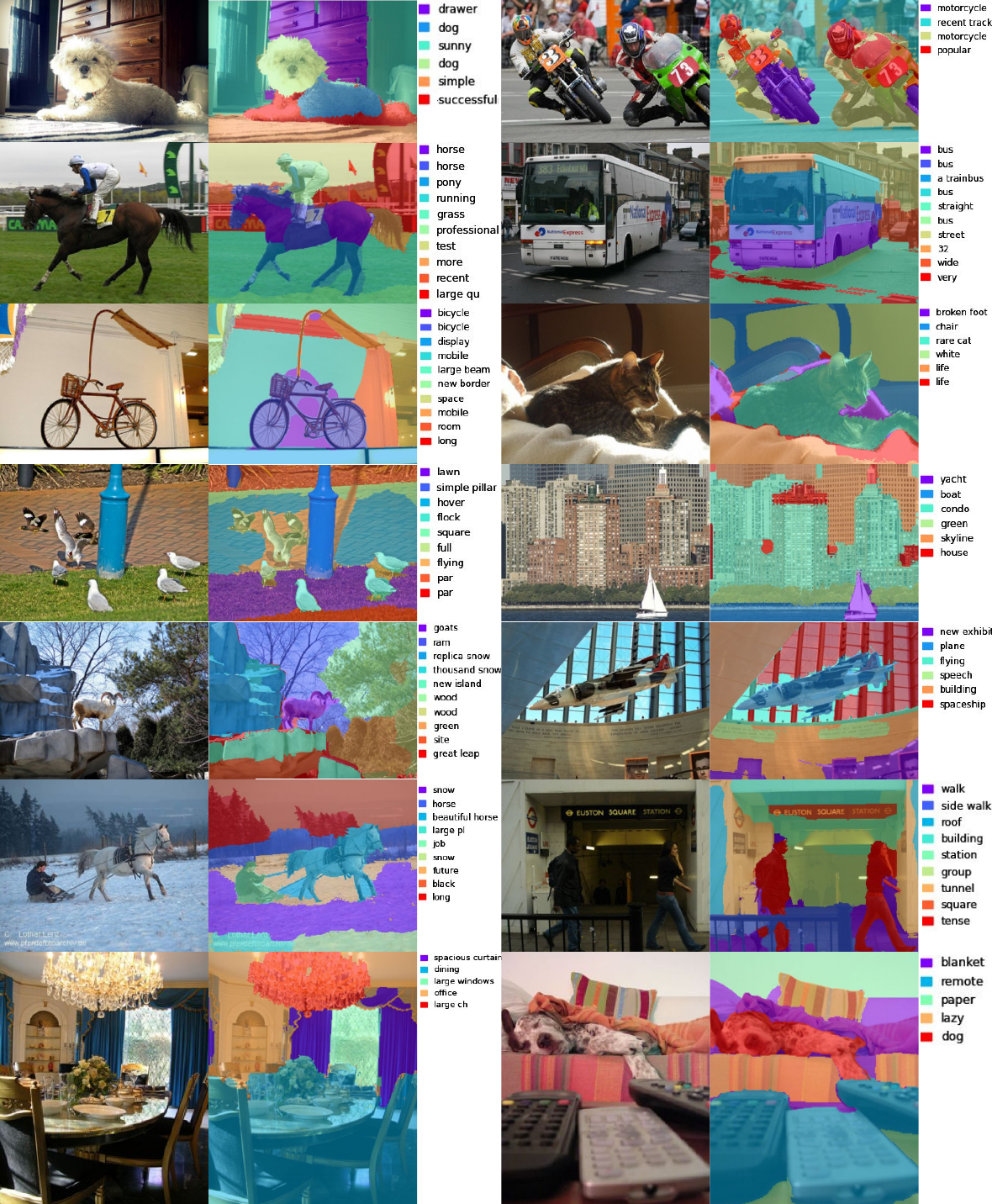}
  \vspace{.5em}
  \caption{Randomly sampled results from Pascal VOC 2012}
  \vspace{8em}
  \label{fig:ran_2012}
\end{figure*}

We present more qualitative results in this section. In Figure \ref{fig:ablation_apx}, we include more results from the ablation experiment in Section 5.3.
We show random results of our method in Figure \ref{fig:ran_context} for the Pascal Context dataset and Figure \ref{fig:ran_2012} for the Pascal VOC 2012 dataset.

\section{Potential Negative Societal Impacts}
\label{apx:negimpact}

Unlike traditional segmentation methods, our method outputs arbitrary text labels and may describe people with incorrect assumptions or discriminatory characteristics based on their stereotypical appearances, such as body shape, clothes, nationality, and sexual orientation. For example, we found `Asian woman' or `homeless' in some generated, which can be offensive in some scenarios. Some characteristics, such as beauty and politics, are rather subjective and challenging to filter without human intervention. Due to the data-driven nature of the pre-trained models we use, our model would also be biased toward the culture, preferences, and characteristics of the training sets and may pose controversial issues.


\clearpage 

\end{document}